# Monitoring a Complex Physical System using a Hybrid Dynamic Bayes Net

**Uri Lerner**
Computer Science Dept.
Stanford University
*uri@cs.stanford.edu*

**Brooks Moses**
Mechanical Engr. Dept.
Stanford University
*bmoses@stanford.edu*

**Maricia Scott**
Computer Science Dept.
Stanford University
*maricia@cs.stanford.edu*

**Sheila McIlraith**
Computer Science Dept.
Stanford University
*sam@ksl.stanford.edu*

**Daphne Koller**
Computer Science Dept.
Stanford University
*koller@cs.stanford.edu*

## Abstract

The *Reverse Water Gas Shift system (RWGS)* is a complex physical system designed to produce oxygen from the carbon dioxide atmosphere on Mars. If sent to Mars, it would operate without human supervision, thus requiring a reliable automated system for monitoring and control. The RWGS presents many challenges typical of real-world systems, including: noisy and biased sensors, nonlinear behavior, effects that are manifested over different time granularities, and unobservability of many important quantities. In this paper we model the RWGS using a hybrid (discrete/continuous) *Dynamic Bayesian Network (DBN)*, where the state at each time slice contains 33 discrete and 184 continuous variables. We show how the system state can be tracked using probabilistic inference over the model. We discuss how to deal with the various challenges presented by the RWGS, providing a suite of techniques that are likely to be useful in a wide range of applications. In particular, we describe a general framework for dealing with nonlinear behavior using numerical integration techniques, extending the successful Unscented Filter. We also show how to use a fixed-point computation to deal with effects that develop at different time scales, specifically rapid changes occurring during slowly changing processes. We test our model using real data collected from the RWGS, demonstrating the feasibility of hybrid DBNs for monitoring complex real-world physical systems.

## 1 Introduction

The *Reverse Water Gas Shift System (RWGS)* shown in Fig. 1 is a complex physical system designed and constructed at NASA's Kennedy Space Center to produce oxygen from carbon dioxide. NASA foresees a number of possible uses for the RWGS, including producing oxygen from the atmosphere on Mars and converting carbon dioxide to oxygen within closed human living quarters.

In a manned Mars mission, the RWGS would operate for 500 or more days without human intervention [Larson and Goodrich, 2000]. This level of autonomy requires the development of robust and adaptive software for fault diagnosis and control. In this paper, we focus on two key subtasks — *monitoring* and *prediction.* Monitoring, or tracking the current state of the system, is a crucial component of the control system. Prediction of the system's expected behavior is a basic tool in fault diagnosis — discrepancies between the predicted and the actual behavior of the system may indicate the presence of faults.

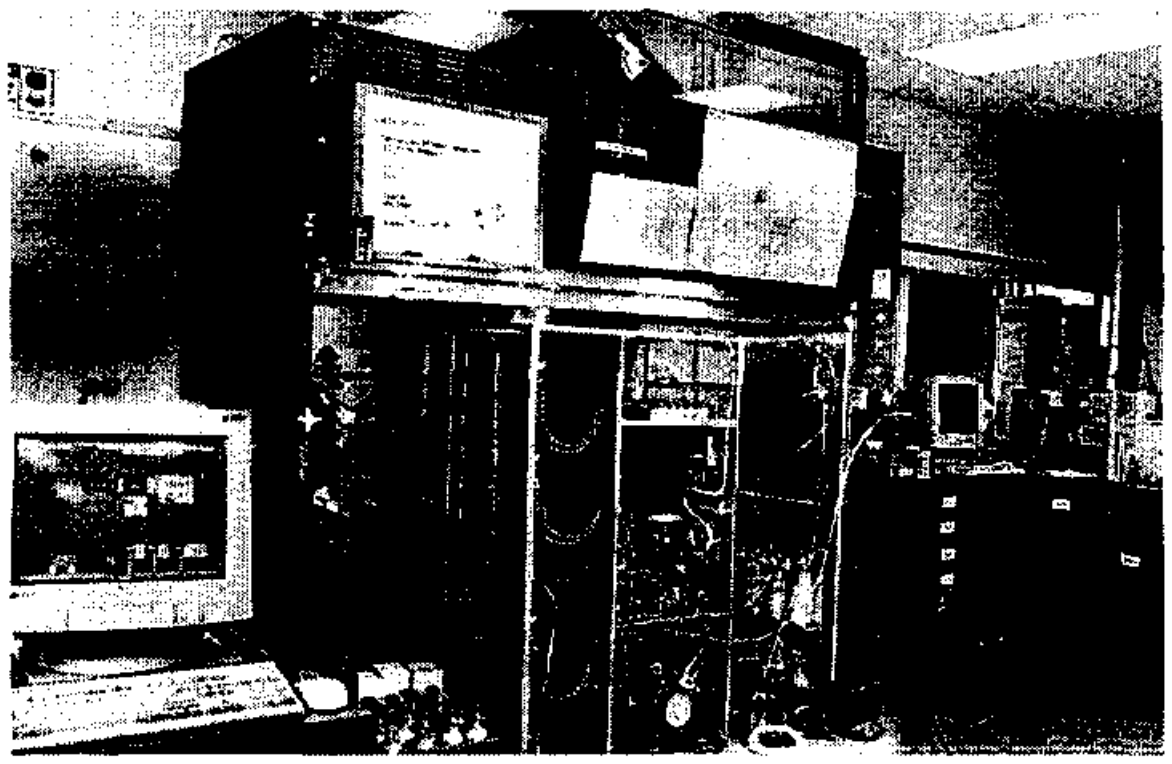

Figure 1: The Prototype RWGS System

The RWGS presents a number of significant modeling and algorithmic challenges. From a modeling perspective, the system is very complex, and contains many subtle phenomena that are difficult to model accurately. Various phenomena in the system manifest themselves over dramatically different time scales, ranging from pressure waves that propagate on a time scale of milliseconds to slow changes such as gas composition that take hours to evolve. From a tracking perspective, the system dynamics are complex and highly nonlinear. Furthermore, the sensors give only a limited view of the system state. Some key quantities of the system are not measured, and the available sensors are noisy and biased, with both the noise level and the bias varying with the system state.

In this paper we model the RWGS using a hybrid (discrete/continuous) *Dynamic Bayesian Network (DBN)*, and show how the system state can be tracked using probabilistic inference over the model. We focus on the continuous part of the model, assuming all the discrete variables are known. We discuss how to deal with the various challenges presented by the RWGS, both in terms of modeling and in terms of inference. We provide a suite of techniques that are likely to be useful in a wide range of applications, in-cluding the case where the discrete variables are not observed.

Perhaps the most interesting modeling problem presented by the RWGS is the issue of different time granularities. A naive solution is to discretize time at the finest granularity. Unfortunately, this approach is generally infeasible both because of the computational burden and because the number of observations is effectively reduced to one for every few thousand time steps, leading to serious inaccuracies. Instead, we take the approach of modeling the system at the time granularity of the observations. We show how to deal with the almost instantaneous changes relative to our time discretization by modeling a part of our system as a set of fixed-point equations.

For the inference task, we provide some new insights into the problem of tracking nonlinear systems. This task is commonly performed using the *Extended Kalman Filter (EKF)* [Bar-Shalom *et al.*, 2001] or the simpler and more accurate *Unscented Filter (UF)* [Julier and Uhlmann, 1997]. We view the problem as a numerical integration problem and demonstrate that the UF is an instance of a numerical integration technique. More importantly, our approach naturally leads to important generalizations of the UF: We show how to take advantage of the structure of the DBN and present a spectrum of filters, trading off accuracy with computational effort.

We tested our model using real data collected from the RWGS prototype system. Our results demonstrate the potential of using hybrid DBNs as a monitoring tool for complex real-world physical systems.

## 2 Preliminaries

In this paper, we characterize physical systems as discrete-time stochastic processes. System behavior is described in terms of a system state which evolves stochastically at discrete time steps $t = 0, 1, 2, \ldots$ We assume that the system is *Markovian* and *stationary*, i.e., the state of the system at time $t + 1$ only depends on its state at time $t$, and the probabilistic dependencies are the same for all $t$.

The system state is modeled by a set of random variables $\mathcal{X} = \{X_1, \ldots, X_n\}$. We partition the state variables $\mathcal{X}$ into a set of evidence (observed) variables, $\boldsymbol{E}$, and a set of hidden (unobserved) variables, $\boldsymbol{H}$. Physical systems commonly comprise both continuous quantities (e.g., flows, pressures, gas compositions) and discrete quantities (e.g., valve open/closed, compressor on/off). Consequently, we model such systems as *hybrid systems*, with $\mathcal{X}$ comprising both discrete and continuous variables.

We model the process dynamics of our system using a *Dynamic Bayesian Network (DBN)* [Dean and Kanazawa, 1989]. A DBN is represented as a Bayes Net fragment called a 2TBN, which defines the *transition model* $P(\boldsymbol{X}' \mid \boldsymbol{X})$ where $\boldsymbol{X}' = \{X_1, \ldots, X_m\}$ denotes the variables at time $t+1$ and $\boldsymbol{X} = \{X_1, \ldots, X_d\}$ denotes some subset of the variables at time $t$ which are *persistent*, in that their values directly influence the next state. More formally, a DBN is a directed acyclic graph, whose nodes are random variables in two consecutive time slices, $\boldsymbol{X}$ and $\boldsymbol{X}'$. The edges in the graph denote direct probabilistic influence between the parents and their child. For every variable $X'$ at time $t + 1$ we denote its parents as $\mathrm{Par}(X') \subset \boldsymbol{X} \cup \boldsymbol{X}'$. Each $X'$ is also annotated with a *Conditional Probability Distribution (CPD)*, that defines the local probability model $P(X' \mid \mathrm{Par}(X'))$. In our hybrid model, discrete nodes do not have continuous nodes as parents.

The *tracking* problem in DBNs is to find the *belief state* distribution$\mathrm{Bel}(\boldsymbol{X}^t) \stackrel{def}{=} P(\boldsymbol{X}^t \mid e^1, ..., e^t)$, where $\boldsymbol{X}^t$ typically consists of the persistent variables $\boldsymbol{X}$ at time $t$, and $e^1, ..., e^t$ are the evidence variables from time 1 to time $t$. The belief state summarizes our beliefs about the state of the system at time $t$, given the observations from time 1 to time $t$. As such, it makes current and future predictions independent of past data. The tracking algorithm is an iterative process that propagates the belief state. We start with the belief state at time $t$, $\mathrm{Bel}(\boldsymbol{X}^t)$ and perform three steps. We first compute $P(\boldsymbol{X}^t, \boldsymbol{X}^{t+1} \mid e^1, ..., e^t)$ as the product $\mathrm{Bel}(\boldsymbol{X}^t)P(\boldsymbol{X}^{t+1} \mid \boldsymbol{X}^t)$. Next we marginalize out $\boldsymbol{X}^t$ resulting in a distribution over $\boldsymbol{X}^{t+1}$. Finally, we condition on $e^{t+1}$, and the result is the belief state at $t + 1$, $\mathrm{Bel}(\boldsymbol{X}^{t+1})$.

*Linear models* are an important class of DBNs. In a linear model, all the variables in $\boldsymbol{X}$ are continuous and all the dependencies are linear with some added Gaussian noise. More precisely, if a node $X$ has parents $Y_1, ..., Y_k$ then $P(X \mid Y_1, ..., Y_k) = \sum_{i=1}^{k} w_i Y_i + V$, where the $w_i$'s are constants and $V$ has a normal distribution $\mathcal{N}(\mu, \sigma^2)$. In a dynamic linear model, tracking can be done using a *Kalman filter* [Kalman, 1960], where the belief state is represented parametrically as a multivariate Gaussian in terms of the mean vector and the covariance matrix. Kalman filters therefore allow a compact belief state representation, which can be propagated in polynomial time and space.

When the dependencies in the model are nonlinear, the resulting distributions are generally non-Gaussian and cannot be represented in closed form. Consequently, the belief state is generally approximated as a multivariate Gaussian that preserves the first two moments of the true distribution. The traditional method for doing this approximation is using an *Extended Kalman Filter (EKF)* [Bar-Shalom *et al.*, 2001]. Assume that $\boldsymbol{X}' = f(\boldsymbol{X})$, where $f$ is some nonlinear function and $\boldsymbol{X} \sim \mathcal{N}(\boldsymbol{\mu}, \Sigma)$. Note that we can always assume that $f$ is deterministic: If the dependency between $\boldsymbol{X}$ and $\boldsymbol{X}'$ is stochastic we can treat the stochasticity as extra random variables that $f$ takes as arguments. The EKF finds a linear approximation to $f$ around the mean of $\boldsymbol{X}$, i.e., we approximate $f$ using the first-order Taylor series expansion around $\boldsymbol{\mu}$. The result is the linear function $f(\boldsymbol{X}) \approx f(\boldsymbol{\mu}) + \nabla f|_{\boldsymbol{\mu}} (\boldsymbol{X} - \boldsymbol{\mu})$, where $\nabla f|_{\boldsymbol{\mu}}$ is the gradient of $f$ evaluated at $\boldsymbol{\mu}$.[1]

[1] A second-order EKF approximation exists, but its increased complexity tends to limit its use.

The EKF has two serious disadvantages. The first is its inaccuracy — the EKF is accurate only if the second and higher-order terms in the Taylor series expansion are negligible. In many practical situations, this is not the case and using the EKF leads to a poor approximation. The second disadvantage is the need to compute the gradient. Some nonlinear functions may not be differentiable (e.g., the max function), preventing the use of an EKF. Even if the function is differentiable, computing the derivatives may be hard if the function is represented as a black box rather than in some analytical form.

The *Unscented Filter (UF)* [Julier and Uhlmann, 1997] provides an alternative approach to tracking nonlinear behavior. As with the EKF, the UF assumes that $\boldsymbol{X}' = f(\boldsymbol{X})$ and $\boldsymbol{X} \sim \mathcal{N}(\boldsymbol{\mu}, \Sigma)$. The UF works by *deterministically* choosing $2d + 1$ points $\boldsymbol{x}_0, ..., \boldsymbol{x}_{2d}$, where $\boldsymbol{x}_0 = \boldsymbol{\mu}$ and the other points are symmetric around $\boldsymbol{\mu}$ (the actual points depend on $\Sigma$). Associated with each point is a weight $w_i$. The UF computes $\boldsymbol{x}'_i = f(\boldsymbol{x}_i)$ for $i = 0, 1, \ldots, 2d$, resulting in $2d + 1$ points in $\mathbb{R}^m$, from which it estimates the first two moments of $\boldsymbol{X}'$ as a weighted average of the $\boldsymbol{x}'_i$'s. In particular, the mean $E[\boldsymbol{X}']$ is approximated as $\sum_{i=0}^{2d} w_i \boldsymbol{x}'_i$.

The UF has several significant advantages over the EKF. First, it is easier to implement and use than the EKF — no derivatives need be computed, and the function $f$ is simply applied to $2d + 1$ points. Second, despite its simplicity, the UF is more accurate than the EKF: The UF is a third-order approximation, i.e., inaccuracies are induced only by terms of degree four or more in the Taylor series expansion. Finally, instead of just ignoring the higher-order terms, the UF can account for some of their effects, by tuning a parameter used in the point selection. As shown in [Julier and Uhlmann, 1997], the UF can be extremely accurate, even in cases where the EKF leads to a poor approximation.

## 3 The RWGS System

The purpose of the RWGS is to decompose carbon dioxide ($CO_2$) (abundant on Mars) into oxygen ($O_2$) and carbon monoxide (CO). The system, shown in Fig. 2(a) [Goodrich, 2002], comprises two loops: a gas loop that converts $CO_2$ and hydrogen ($H_2$) into $H_2O$ and CO, and a water loop that electrolyzes the $H_2O$ to produce $O_2$ and $H_2$. Under normal operation, $CO_2$ at line (1) is combined with $H_2$ returned from the electrolyzer via line (12), and a mixture of $CO_2$, $H_2$, and CO from the reactor recycle line (11). This mixture enters a catalyzed reactor (3) heated to 400°C. Approximately 10% of the $CO_2$ and $H_2$ react to form CO and $H_2O$:

$$CO_2 + H_2 \rightleftharpoons CO + H_2O$$

The $H_2O$ is condensed at (4) and is stored in a tank (5). The remaining gas mixture passes through a separation membrane (9), which sends a fraction of the CO to the vent (13) while directing the remaining mixture into the recycle line (11). A compressor (10) is used to maintain the necessary pressure differential across the membrane. In the water loop, the $H_2O$ in tank (5) has some $CO_2$ dissolved in it, which would be detrimental to the electrolyzation process. To remedy this, the $H_2O$ is pumped into a second tank (6), and has $H_2$ bubbled through it to purge the $CO_2$. From there, the $H_2O$ is pumped into the electrolyzer (8), which separates a portion of it into $O_2$ and $H_2$. The $H_2$ re-enters the gas loop via (12), while the remaining $H_2O$, along with the $O_2$, goes into tank (7), where the mixture is cooled and separated. The $H_2O$ returns to the electrolyzer, while the $O_2$ leaves the system through (14).

In addition to its normal operating mode, the system may operate without the electrolyzer and water pumps. In this mode, the $H_2$ for the reaction is supplied by a supply line (15) paralleling the $CO_2$ supply line. This option is not feasible for operation on Mars, but has proven useful for testing the physical system while under development.

The RWGS is an interconnected nonlinear system where the various components influence each other in complicated and sometimes unexpected ways. For example, during runs without the electrolyzer, it is necessary to empty the water tank (5) periodically, to prevent water from accumulating and eventually overflowing the tank. This causes the gases in the tank to expand, and thus creates a significant and sudden pressure drop, which affects the flow throughout the whole system. This phenomenon is demonstrated in Fig. 2(b), taken from [Whitlow, 2001]. The graph shows the flow through the CO vent (13) as it evolves over time — the spikes correspond to emptying the water tank.

A challenging property of the RWGS is that phenomena in the system manifest themselves over at least three different time scales. Pressure waves in the RWGS propagate essentially instantaneously (at the speed of sound). Gases flow around the gas loop on the order of seconds. Finally, gas compositions in the gas loop take on the order of hours to reach a steady state. Meanwhile, the sensors collect data at a sampling rate of one second.

An additional challenge of the RWGS is its sensitivity and unidentifiability, i.e., parts of the system state are very sensitive to input paramaters and are not directly measured. For example, the $H_2$ and $CO_2$ compositions in the gas loop cannot be practically measured. However, the balance between these compositions is almost neutrally stable, thus a small shift in the input conditions or the membrane behavior will cause the balance to gradually drift to a significantly different value.

As in any real system, the RWGS sensors do not record the underlying state exactly. In addition to some important quantities, such as the gas compositions, which are not measured at all, the existing sensors are noisy and biased. The noise level of the sensors depends on many factors and can change over time. An example is shown in Fig. 2(c), where the difference in the readings of the pressure sensors $P_3$ and $P_4$ (both located at (2) in Fig. 2(a)) is plotted over time. The main reason for the noise in time steps 0–800 is the physical proximity of the sensors to the compressor that sends pressure waves throughout the system. Since the sensors are not synchronized with the compressor, they take measurements at various phases of the pressure waves and thus measure significantly different values. After 796 seconds the compressor shuts down and the noise level decreases dramatically. [2] More interestingly, we note that the two sensors are placed very close together and thus the average difference should be zero. However, as the plot demonstrates, this is not the case, indicating that the sensors are not well calibrated and some bias is present. Furthermore this bias depends on the system state, as shown by the change in the average difference when the compressor shuts off.

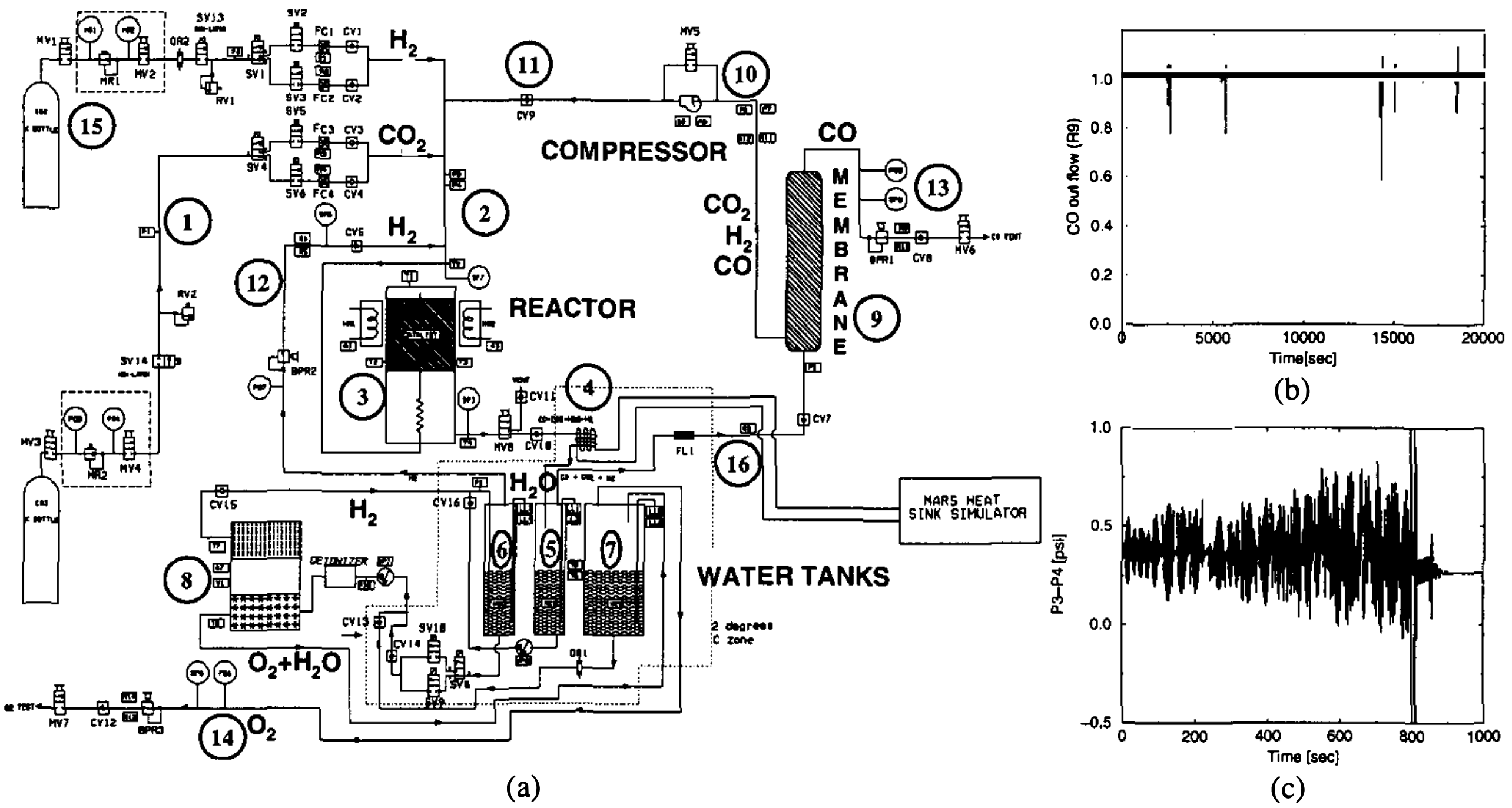

Figure 2: (a) The RWGS Schematic. (b) Effects of emptying a water tank. (c) Pressure difference between $P_3$ and $P_4$.

## 4 Modeling the RWGS

We model the RWGS using a hybrid DBN, as described in Section 2. The 2TBN has 293 nodes, 227 of which are continuous. Currently the discrete variables in the model are all known and correspond to computer-controlled switches and sensor faults. The continuous variables in our model capture the continuous-valued elements of our system (e.g., pressure at various points in the system, flow rates, temperatures, gas composition, etc.). Of the 227 continuous nodes, 43 represent the time $t$ belief state $\boldsymbol{X}$ and 184 represent the variables $\boldsymbol{X}'$ at time $t{+}1$. Of the latter, 43 variables are belief state variables for $t+1$, 72 variables are *encapsulated* variables, as discussed in Section 5.4, and the rest are either sensor variables or transient variables.

When constructing the model, we used four techniques for parameter estimation. Some of the parameters were known physical constants or system properties. Of the empirical parameters, many came from physical models. The others (specifically, some parameters for the compressor, the separation membrane and the overall system pressure changes) were determined using common equations that model the particular system behavior. All the variables in these equations were directly observed in the data, and thus we could use least-squares techniques to find the best fit for the parameters. The remaining parameters were estimated using prior knowledge of the domain.

### 4.1 Sensor Modeling

As discussed in Section 3, one of the challenges we address in modeling the RWGS is dealing with noisy and biased sensors. We deal with noisy sensors in the obvious way: by increasing the variance of the predicted measurement values to match the noise level in the data.

Sensor biases present a more interesting modeling problem. The biases are not easily modeled using a simple parameter since they are unknown and can drift over time. Instead, we address the problem by adding hidden variables to the belief state that model the different biases of the sensors. Biases start with zero mean and a reasonably large variance and persist over time, i.e., $\text{Bias}^{t+1} = \text{Bias}^{t} + V$, where $V$ represents white noise with a relatively small variance, allowing for some amount of drift to occur over time.

This idea works quite well, but it tends to overfit the data: By letting the bias account for every discrepancy between the model predictions and the actual sensor measurements, the tracking algorithm might settle in an incorrect steady state. To fix the problem we must make sure that the model biases reflect true sensor biases — biases should be kept as small as possible and allowed to grow only if there is a real reason for that. We implement this idea by introducing a contraction factor $\gamma < 1$ (empirically set to be $0.97$) into the bias formula: $\text{Bias}^{t+1} = \gamma \cdot \text{Bias}^{t} + V$. Thus, biases tend to go to zero unless doing so introduces a systematic discrepancy with the predicted system state.

[2] The sensor's noise is literally noise that can be heard — the pressure waves are the sound waves generated by the compressor.

### 4.2 Sensitivity and Unidentifiability

Recall that the equations governing the $H_2/CO_2$ balance in the gas loop are sensitive to slight variations in the physical parameters. Thus even using the most exact form of these equations in the model will result in (at least) the same level of sensitivity — both to variations in the physical parameters, and inherent errors in the parameters. Moreover, the model value is also sensitive to model effects such as calculation errors and sensor errors that do not affect the real value. We therefore use equations for the $H_2/CO_2$ balance that contain an intentionally non-physical component—a *stabilizing* term—that reduces the sensitivity. This term drives the balance to a pre-determined point, which in this case is our expected value for the balance. The magnitude of this term is manually adjusted to provide an optimum tradeoff between physical accuracy and model stability.

### 4.3 Differing Time Scales

As described in Section 3, we must deal with differing time scales in modeling the RWGS. The naive solution to this problem is to model the DBN at a very fine time granularity. However, it is completely impractical to model the behavior of the pressure waves using a discretized-time model. To do so would require time steps three orders of magnitude smaller than the time between measurements, which is a significant waste of resources. Furthermore, it would require a much more complete description of the system than is practical, and tracking the slowly-evolving aspects of the system with a step size many orders of magnitude below their time scale would allow substantial errors to build up.

Thus, we approximate the pressure waves as occurring instantaneously and instead of modeling their transient behavior, we model the quasi-steady-state results at each time step after they have reached an equilibrium. The equations in this case are substantially simpler, and require far fewer empirical constants. The difficulty, however, is that these equations must be solved simultaneously; a change in any part of the system will affect all of the other parts. These equations include both the compressor equation and an approximation to the membrane equations developed in [Whitlow, 2001]; thus, they are fairly large and nonlinear, and no direct simultaneous solution form exists. Instead, we use these equations to create a new equation that converges to a fixed point solution.

We must insert this fixed-point equation into a (nonlinear) CPD to use it in our DBN model of the RWGS. The equation solves for the five model variables $\boldsymbol{Z}$ that account for the flows and pressure of the gas loop. In order to solve for all five variables, their eight parents must also be present in the CPD. Hence, we have a vector CPD for $\boldsymbol{Z}$ whose definition is essentially procedural: given a value of the eight parents it executes an iterative fixed-point computation until convergence, and outputs the values $\boldsymbol{Z}$.

## 5 Tracking in Nonlinear Systems

In this section, we address the problem of inference, focusing on tracking in complex nonlinear systems, such as the RWGS. In these models, the probabilistic dependencies, including sensors, can be either linear or nonlinear functions with Gaussian noise. We restrict our attention to the task of tracking the continuous state, assuming all the discrete values are known. Note that although the results in this section are presented in terms of dynamical systems, the analysis also applies to probabilistic inference in *static* nonlinear Bayes nets.

### 5.1 Exploiting DBN Structure

Recall the setup from Section 2: We have a Gaussian belief state $\text{Bel}(\boldsymbol{X})$ where $\boldsymbol{X} \in \mathbb{R}^d$ and a 2TBN representing $P(\boldsymbol{X}' \mid \boldsymbol{X})$ as a deterministic function $\boldsymbol{X}' = f(\boldsymbol{X})$. Our goal is to find an approximation of $P(\boldsymbol{X}')$ as a multivariate Gaussian. The classical approach, used in the EKF and the UF, is to find the entire distribution $P(\boldsymbol{X}')$ directly by treating $f$ as a function from $\mathbb{R}^d$ to $\mathbb{R}^m$. An alternative approach is to decompose $f$ by defining $X_i' = f_i(\boldsymbol{Y}_i)$ for $i = 1, \ldots, m$, where $\boldsymbol{Y}_i = \text{Par}(X_i')$. In most practical cases the $f_i$'s have a lower dimension than $f$; as we shall see, this reduction in the dimension lets us approximate the resulting distribution more accurately and efficiently.

As discussed in Section 2, the first step in the belief state propagation process is to compute a multivariate Gaussian over $\{\boldsymbol{X}, \boldsymbol{X}'\}$. We begin with our Gaussian $\text{Bel}(\boldsymbol{X})$, and add the variables from $\boldsymbol{X}'$ one at a time, using the procedure described in Section 5.2. The key insight is that, as $X_i'$ is conditionally independent of $\{\boldsymbol{X} - \boldsymbol{Y}_i, X_1', \ldots, X_{i-1}'\}$ given $\boldsymbol{Y}_i$, it suffices to approximate the Gaussian $P(\boldsymbol{Y}_i, X_i')$. We can then compute $P(\boldsymbol{X}, X_1', \ldots, X_i') = P(\boldsymbol{X}, X_1', \ldots, X_{i-1}') P(X_i' \mid \boldsymbol{Y}_i)$, which, for Gaussians, can be accomplished using simple linear algebra operations.

A more difficult case arises when the DBN contains not only inter-temporal edges from $\boldsymbol{X}$ to $\boldsymbol{X}'$, but also intra-temporal edges between $\boldsymbol{X}'$ variables. In this case we sort the variables $X_i'$ in topological order, and gradually build up the joint distribution $P(\boldsymbol{X}, X_1', \ldots, X_i')$. The topological order ensures that when we need to compute $P(\boldsymbol{Y}_i, X_i')$, we have already computed a Gaussian over $\boldsymbol{Y}_i \subseteq \boldsymbol{X} \cup \{X_1', \ldots, X_{i-1}'\}$. This approach, however, may introduce some new inaccuracies, because we now also use a Gaussian approximation for the distribution of the relevant variables from $\{X_1', \ldots, X_{i-1}'\}$.

Even in cases where we introduce extra inaccuracies, this method is often superior to the UF. The reason is that, by reducing the dimension of the functions involved, we can use more accurate techniques to approximate the first two moments of the variables in $\boldsymbol{X}'$ with the same computational resources. In general, there is a tradeoff between the superior precision we achieve for each variable and the potential for extra inaccuracies we introduce. The extra inaccuracies depend on the quality of our Gaussian approximation for $P(\boldsymbol{X}, X'_1, \ldots, X'_{i-1})$, and on the extent of the nonlinearity of the dependencies within $\boldsymbol{X}'$. If the dependence of $X'_i$ on $\{X'_1, \ldots, X'_{i-1}\}$ is linear, then there are no extra errors introduced: In this case the first two moments of $X'_i$ are only influenced by the first two moments of $\{X'_1, \ldots, X'_{i-1}\}$ which can be captured correctly by our Gaussian approximation. It is somewhat reassuring that the better our approximation of $P(\boldsymbol{X}')$ as a Gaussian is, the less significant the extra errors we introduce are, as the entire framework is based on the assumption that $P(\boldsymbol{X}')$ can be well approximated by a Gaussian.

### 5.2 Numerical Integration

We now turn our attention to the task of approximating $P(\boldsymbol{Y}_i, X'_i)$ as a multivariate Gaussian. To simplify our notation, let $X$ be a variable which is a nonlinear function of its parents $\boldsymbol{Y} = Y_1, \ldots, Y_d$, i.e., $X = f(\boldsymbol{Y})$, but the ensuing discussion also holds for the vector case of $\boldsymbol{X} = f(\boldsymbol{Y})$. We assume that $P(\boldsymbol{Y})$ is a known multivariate Gaussian, and the goal is to find a Gaussian approximation for $P(\boldsymbol{Y}, X)$. It suffices to compute the first two moments:

$$E[X] = \int P(\boldsymbol{Y})f(\boldsymbol{Y})d\boldsymbol{Y} \quad (1)$$

$$E[X^2] = \int P(\boldsymbol{Y})f^2(\boldsymbol{Y})d\boldsymbol{Y} \quad (2)$$

$$E[XY_j] = \int P(\boldsymbol{Y})f(\boldsymbol{Y})Y_j d\boldsymbol{Y} \quad (3)$$

Note that the integrals only involve the direct parents of $X$, significantly reducing their dimension. We can effectively compute these integrals using a version of the Gaussian Quadrature method called the *Exact Monomial* rules [Davis and Rabinovitz, 1984]. Generally speaking, Gaussian Quadrature approximates integrals using a formula of the form:

$$\int W(\boldsymbol{x})f(\boldsymbol{x})d\boldsymbol{x} \approx \sum_{j=1}^{N} w_j f(\boldsymbol{x}_j)$$

where $W(\boldsymbol{x})$ is a known function (a Gaussian in our case). The points $\boldsymbol{x}_j$ and weights $w_j$ are carefully chosen to ensure that this approximation is exact for any polynomial $f$ whose degree is at most $p$. The degree $p$ is called the *precision* of the approximation.

Finding a set of points with a minimal size $N$ for some precision $p$ is not a trivial task. In the simple form of Gaussian Quadrature, we choose points in one dimension and use them to create a grid of points in $\mathbb{R}^d$ with the obvious disadvantage that $N$ grows exponentially with $d$. Fortunately, we can do better. In [McNamee and Stenger, 1967] a general method is presented for $N = O\left(\frac{(2d)^k}{k!}\right)$ and precision $p = 2k + 1$ ($d$ is the dimension of the integral, in our case $|\boldsymbol{Y}|$). In particular, rules are presented for $2d + 1$ points with precision 3, $2d^2 + 1$ points with precision 5 and $\frac{4}{3}d^3 + \frac{8}{3}d + 1$ points with precision 7. The precision 3 rule is exactly the rule used for the Unscented Filter: It has exactly the same $2d + 1$ points and weights.

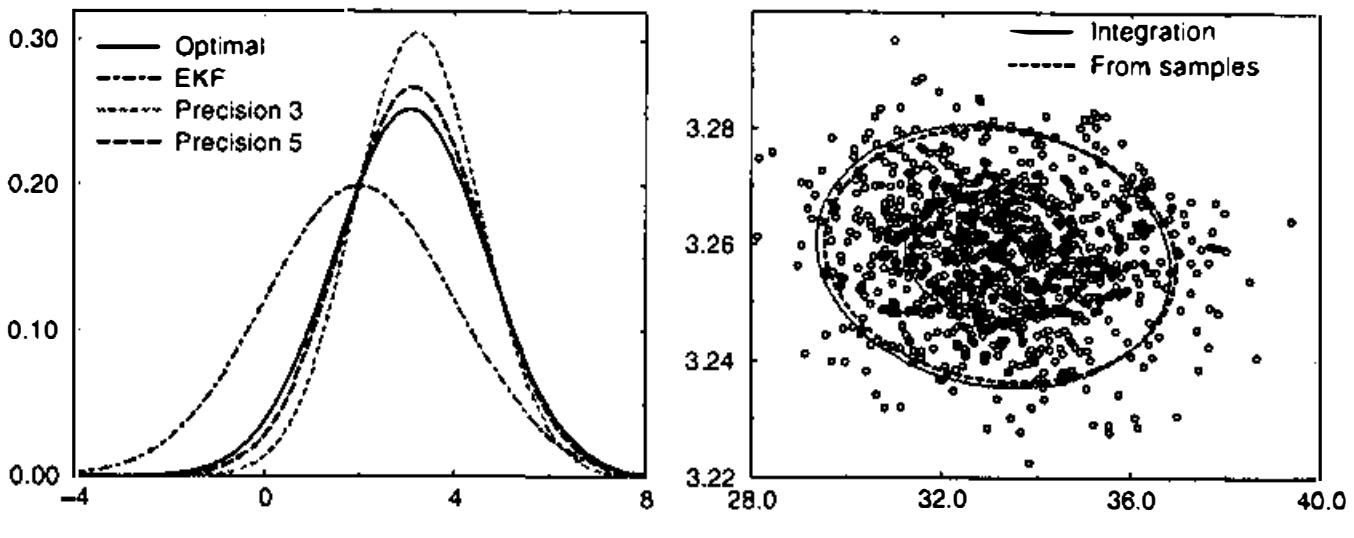

(a) (b)

Figure 3: (a) Density estimates for $X = \sqrt{Y_1^2 + Y_2^2}$. (b) Random samples from the RWGS network for the flow at point (16) and the pressure at point (2), and Gaussian estimates for the distribution.

This view of the Unscented Filter has immediate practical consequences: we can trade off between the accuracy of the computation and its computational requirements. For example, if we are interested in a more precise filter than the Unscented Filter and are willing to evaluate the function at $O(d^2)$ points then we can use the exact monomial rule of precision 5. Depending on the function, this may represent a significant gain in accuracy.

As a simple example we consider the nonlinear function $X = \sqrt{Y_1^2 + Y_2^2}$ where $P(Y_1) = \mathcal{N}(2, 4)$ and $P(Y_2 \mid Y_1) = \mathcal{N}(0.5Y_1 - 1, 3)$ (note that both $Y_1$ and $Y_2$ have the same variance 4). Fig. 3(a) shows various estimates for the probability of $X$. The optimal estimate is the best Gaussian approximation for the distribution of $X$ computed using a very exact numerical integration rule. We can see that the exact monomial rules of precisions 3 and 5 provide a much better estimate than EKF, where the precision 5 rule leads to a more accurate estimate than the precision 3 rule.

### 5.3 Inaccuracies in the Approximation

Unfortunately, approximating $P(\boldsymbol{Y}, X)$ using numerical integration can lead to covariance matrices that are not semi-positive definite, and hence illegal. One simple approach to this problem is to use a more accurate integration rule, although the problem may persist. An alternative is to find the "closest" positive definite covariance matrix. We cast this problem as a convex optimization problem following [Boyd and Vandenberghe, 2003].

Consider once again the problem of approximating $P(\boldsymbol{Y}, X)$ as a multivariate Gaussian, where $X$ is a nonlinear function of its parents $\boldsymbol{Y}$, i.e., $X = f(\boldsymbol{Y})$, and $\boldsymbol{Y} \sim \mathcal{N}(\boldsymbol{\mu_Y}, \Sigma_{\boldsymbol{YY}})$. Let $\Sigma$ denote the estimated covari-ance matrix for $P(\boldsymbol{Y}, X)$:

$$\Sigma = \begin{pmatrix} \Sigma_{\boldsymbol{YY}} & \bar{\boldsymbol{u}} \\ \bar{\boldsymbol{u}}^T & \bar{v} \end{pmatrix}$$

If $\bar{\boldsymbol{u}}$ and $\bar{v}$ lead to a matrix $\Sigma$ that is not positive definite, then we need to find the closest $\boldsymbol{u}$ and $v$ to $\bar{\boldsymbol{u}}$ and $\bar{v}$, such that $\Sigma$ is positive definite. Given that $\Sigma_{\boldsymbol{YY}}$ is already positive definite, $\Sigma$ is positive definite iff $v - \boldsymbol{u}^T \Sigma_{\boldsymbol{YY}}^{-1} \boldsymbol{u} > 0$. Thus, we can formalize our problem as follows:

$$\text{Minimize} \quad \| \boldsymbol{u} - \bar{\boldsymbol{u}} \|^2 + (v - \bar{v})^2 \qquad (4)$$

$$\text{Subject to} \quad \boldsymbol{u}^T \Sigma_{\boldsymbol{YY}}^{-1} \boldsymbol{u} - v + \epsilon \leq 0 \qquad (5)$$

where $\epsilon$ is some small positive number. Since both Eq. 4 and Eq. 5 are convex we can solve this problem by forming the Lagrangian and solving the dual problem. We set the partial derivatives of $\boldsymbol{u}$ and $v$ to zero and plug the result into Eq. 5. We get an equation over the Lagrangian multiplier which can be solved easily as it involves a monotonic function. We omit details for lack of space.

Our analysis treats the elements in $\boldsymbol{u}$ and $v$ directly, but in fact these elements are not independent since $u_i = E[Y_i X] - \mu_{Y_i} E[X]$ and $v = E[X^2] - E[X]^2$. It is desirable to use this relation in Eq. 4 and Eq. 5 and represent the dependency between the various elements (e.g., a change in $E[X]$ may fix many of the problems simultaneously). Unfortunately, because of the term $E[X]^2$ the problem is no longer convex. Nonetheless, we can approximate the problem as convex (e.g., by replacing $E[X]^2$ by the best current estimate), solve it and iterate. Again, we defer details to an extended version of this paper.

### 5.4 Encapsulated Variables

Just as we can use the DBN structure to decompose the dependency between $\boldsymbol{X}'$ and $\boldsymbol{X}$, in many cases we can further decompose the dependency $X = f(\boldsymbol{Y})$. For example, assume that $f(\boldsymbol{Y}) = g(g_1(\boldsymbol{Y}_1), g_2(\boldsymbol{Y}_2))$, where $\boldsymbol{Y}_1, \boldsymbol{Y}_2 \subseteq \boldsymbol{Y}$.[3] Instead of directly evaluating the Gaussian over $\{\boldsymbol{Y}, X\}$ we can define two extra variables: $T_1 = g_1(\boldsymbol{Y}_1)$ and $T_2 = g_2(\boldsymbol{Y}_2)$. We first approximate $P(\boldsymbol{Y}_1, T_1)$ as a Gaussian and use it to find a Gaussian over $\{\boldsymbol{Y}, T_1\}$. Next we approximate $P(\boldsymbol{Y}_2, T_2)$ as a Gaussian and from it $P(\boldsymbol{Y}, T_1, T_2)$. Finally, we approximate $P(T_1, T_2, X)$ as a Gaussian and use it to find the Gaussian approximation for $P(\boldsymbol{Y}, T_1, T_2, X)$. The same accuracy tradeoffs that were discussed in the context of $\boldsymbol{X}' = f(\boldsymbol{X})$ apply here: by reducing the dimension of the integrals we can solve each one more accurately, but may introduce further errors if the interaction between the extra variables is nonlinear.

[3]E.g., flow sensors give different results depending on the gas type. Assuming we have random variables representing the total flow and the compositions of the different gases in it, $g_1$ and $g_2$ may each be a product of one of the gas compositions and the flow, thus representing the net flow of a certain gas. The function $g$ would be a weighted sum of these flows where the weights correspond to the sensor's response for the different gases.

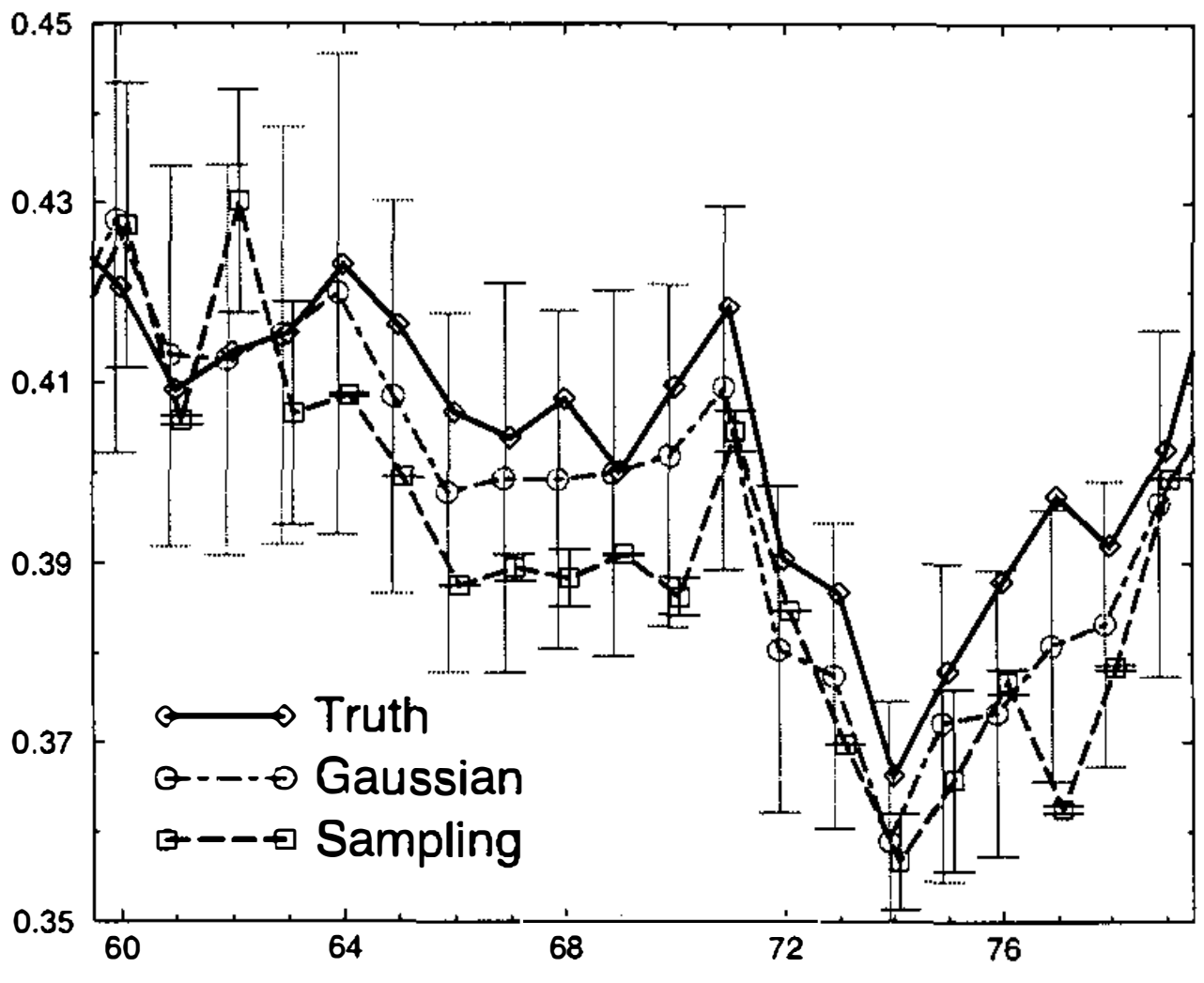

Figure 4: Comparison with particle filtering on simulated data, showing the means and error bars of two standard deviations for our algorithm and particle filtering. The $X$ axis represents time, and the $Y$ axis the percentage of $H_2$ in the flow at point (16). To increase readability, we shift the estimates generated by our algorithm by 0.1 to the left and those generated by particle filtering by 0.1 to the right.

In principle, one could add $T_1$ and $T_2$ to the DBN and treat them as regular variables. However, doing so makes these variables part of $\boldsymbol{X}'$, and thereby increases the algorithm's space complexity, which is $O(|\boldsymbol{X}'|^2)$ (for representing the covariance matrix of $P(\boldsymbol{X}')$). It is better to treat the extra variables as local variables *encapsulated* within the CPD and unknown to the rest of the network. After computing the Gaussian approximation for the CPD variables, we simply marginalize over the encapsulated ones. This approach is similar to the local computations in an OOBN model [Koller and Pfeffer, 1997], where some of the CPD variables are encapsulated within the CPD.

## 6 Experimental Results

In this section we present results from a set of experiments that test the efficacy and robustness of our model and tracking algorithm. Our computational model of the RWGS contains all of the components needed to monitor the full operation of the physical system, although data provided to date by KSC is for the reduced-operation mode with only the gas loop component operational. Our experiments were run on a Pentium III 700MHz.

We tested our algorithm with both real data and simulated data that was generated from our model. Although running with real data is the real test for our approach, running with simulated data is also of interest. The reason is that there are two sources of errors when using real data: model inaccuracies and errors induced by the algorithm. When using simulated data, only errors of the second type are present and we can better test the performance of the algorithm.

### 6.1 Results on Simulated Data

We first tested whether the belief state could be well approximated as a Gaussian and whether our particular approximation was a good one. To do so, we generated a set of samples from the model. We did not introduce any evidence so the samples were indeed sampled from the correct joint distribution. In Fig. 3(b) we show the results for two particular variables: the flow at point (16) and the pressure at point (2) (these variables were chosen because of their dependency on the non-linear CPD of the membrane; other variables produced similar results). The samples appear to be drawn from a distribution that is either a Gaussian or close to one. Furthermore, our estimate for the joint distribution (depicted by the contours for one and two standard deviations) is very close to the Gaussian that was estimated directly from the samples. Thus, it is reasonable to expect that our techniques will lead to good approximations of the belief state.

Next, we generated a trajectory of 500 time steps from our model and tested our algorithm on it. We compared our results with the particle filtering algorithm [Gordon *et al.*, 1993], which approximates the belief state as a set of weighted samples where the weights of the samples correspond to the likelihood of the evidence given the sample. Our algorithm took 20ms per time step, which included computing the Gaussian approximation to the belief state, with numerical integration when necessary, and conditioning on the evidence. In comparison, generating a sample using particle filtering took 1.5ms. Thus, one step of our algorithm took as much time as generating 13 samples. However, with just 13 samples particle filtering performed extremely poorly and therefore in our experiments we used 10,000 samples at every time step, giving particle filtering a somewhat unfair advantage.

Fig. 4 shows the estimates for the percentage of $H_2$ in the flow at point (16) that were computed by our algorithm and by particle filtering, as well as the actual value (known from the simulated data). We report the results on this particular variable because the gas compositions are not measured by any sensors and are therefore a potential challenge to our algorithm. The error bars represent the uncertainty of the estimates as plus and minus two standard deviations (for particle filtering we computed the standard deviation induced by the weighted samples).

Although under our setup particle filtering was slower than our algorithm by a factor of 750, as Fig. 4 demonstrates, the estimates of particle filtering are not as good as the estimates of our algorithm. Over the entire sequence the average error of our algorithm was $0.009$ while the average error of particle filtering was $0.013$. Nevertheless, the more dramatic difference is in the estimates of the variance. Often, the estimated variance for particle filtering is extremely small, even when the estimated value is not very accurate (e.g., time steps 72 and 73). In fact, over the entire sequence, according to the estimated distribution of our algorithm, the correct value of the $H_2$ composition was within two standard deviations 96% of the time (this is consistent with the fact that the probability mass within two standard deviations from a Gaussian mean is 95%). In comparison, for particle filtering, the true value was within two estimated standard deviations only 20% of the time. The difference was even more apparent when we computed the average log-likelihood of the true value, given the two possible estimates. For our algorithm the average log-likelihood was $3.1$ while for particle filtering it was only $-5.59 \cdot 10^{11}$.

The reason for this problem is the relatively high dimension of the evidence which leads to a very high variance for the weights of the samples. Although we generated 10,000 samples at each time step only a very small number of them had a significant effect on the estimate. Over the entire sequence, in 65% of the time steps one sample accounted for more than 0.5 of the total probability mass, in 27% one sample accounted for more than 0.9 of the mass, and in 15% one sample accounted for more than 0.99. Obviously in cases where one sample completely dominates the rest, the estimates of particle filtering are not very reliable and in particular the variance estimates can be extremely small and misleading.

Thus, not only is our algorithm faster than particle filtering with 10,000 samples by a factor of 750, its estimates are much more reliable.

### 6.2 Results on Real Data

We next ran a set of experiments on real data. Our data set consisted of a long sequence of 13,875 time steps, most of it collected while the system was running in steady state. We divided our data into a training set, used to estimate and tune the model parameters, and a test set on which we report our results.

We conducted a variety of experiments in which we compared model predictions with the actual measurements recorded by the system under various scenarios: steady state and non-steady state, removing sensors, and modifying the sensor models. In order to make the comparison informative, the model predictions for values at time $t+1$ as reported in this section are not adjusted with evidence at time $t+1$, i.e., they are the predictions based on evidence from times $0, 1, \ldots, t$.

Our first experiment, shown in Fig. 5(a), illustrates the efficacy of our tracking algorithm during steady-state operation of the system. In particular, the graph illustrates the predicted (thick lines) and measured (thin lines) pressures, $P_3$ and $P_4$ at point (2) in Fig. 2(a). Observe that the predicted value for $P_3$ appears to be consistently lower than the measurement. This is the result of the model's bias weighting, $\gamma = .97$, discussed in Section 4.1, which

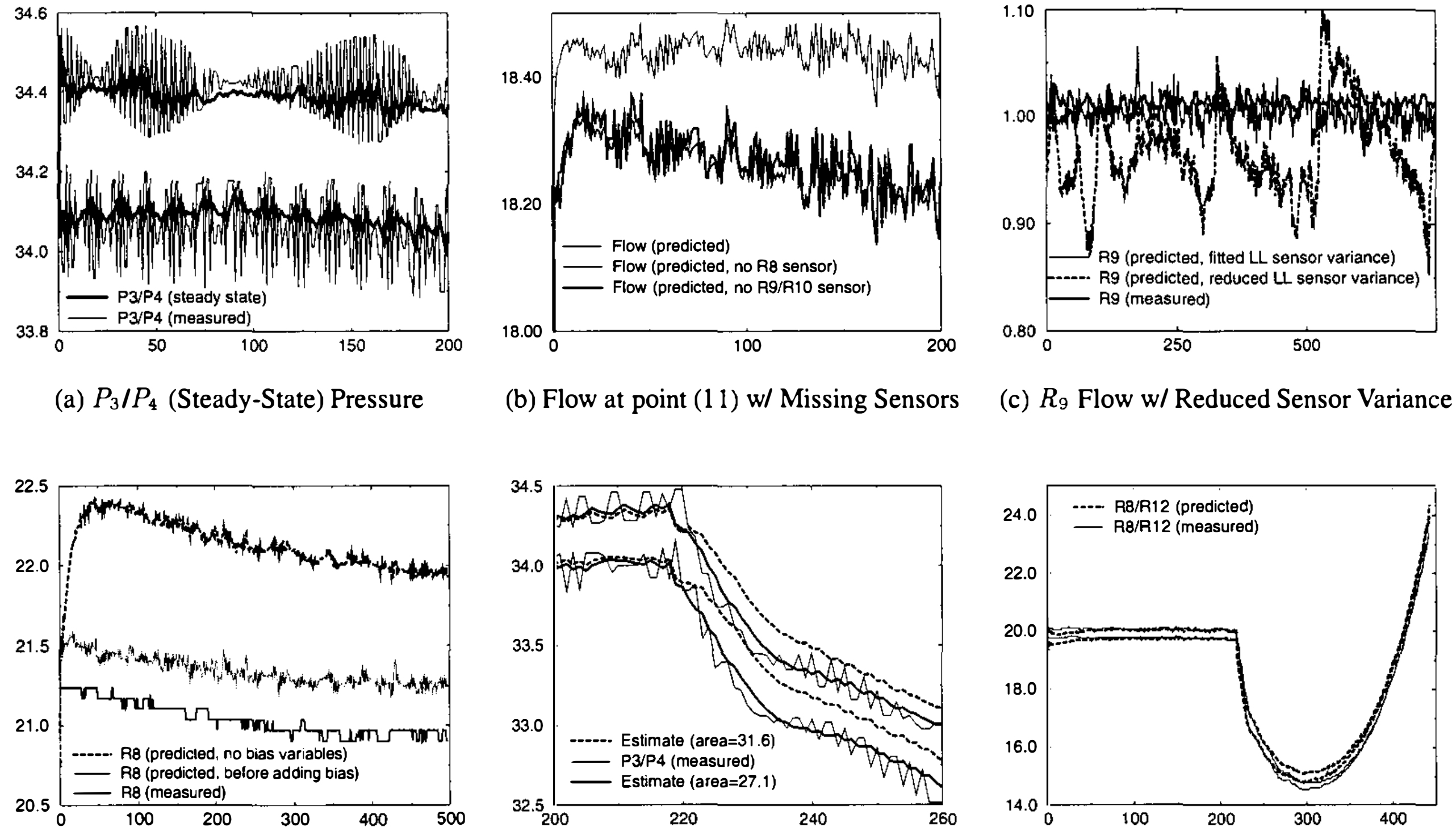

(a) $P_3/P_4$ (Steady-State) Pressure (b) Flow at point (11) w/ Missing Sensors (c) $R_9$ Flow w/ Reduced Sensor Variance

(d) $R_8$ Flow with(out) Bias (e) Pressure Transition (f) Flow Transition

Figure 5: Experimental Results Tracking the RWGS. The $X$ axis represents time, and the $Y$ axis the value of the appropriate quantity.

tends to pull the estimates slightly away from the measured value. While, in this case, it produces a slightly poorer result, overall, the bias weighting technique does less data overfitting and works better in non-steady state sequences.

Next we experimented with "removing" sensors from the system. (This is easily achieved by ignoring selected sensor evidence when running the tracking algorithm.) Sensor removal can be used to evaluate the robustness of the algorithm as well as to determine the importance of a sensor for monitoring the system. In Fig. 5(b), we show the flow of gas from the compressor at point (11). The two overlaid lines are our estimates of this flow value — one with all of the sensors, and the other with sensors $R_9$ and $R_{10}$ (located at (13)) removed. In contrast, when flow sensor $R_8$ (located at (16)) is removed, the predicted flow rate quickly strays. These results indicate that, at least for this sequence, $R_8$ is a more valuable sensor than $R_9$ and $R_{10}$.

We also tested the effects of changing the liquid level (LL) sensor noise parameter [4] on our prediction of the gas flow $R_9$ at (13). Recall from Section 4.1 that to correctly model a sensor we introduced both some Gaussian noise on the sensor and a hidden bias variable. We tried both a variance value of 0.01, which we estimated using "reasonable" prior knowledge, and a variance value of 4 which was fit to the data. Fig. 5(c) shows the effect of the variance of the LL sensor for the water tank at (5). With the fitted variance, the algorithm tracked quite well. In contrast, with the smaller variance, the performance was poor and erratic, following the fluctuations in the LL measurements.

The utility of the bias variables is shown in Fig. 5(d). The upper line is a prediction of the flow rate, made using a version of the model that contained no bias variables for the flow sensors at (10), (13) and (16). The middle line corresponds to the model with the bias variables present, but shows the prediction for the true (unbiased) flow (i.e., the sensor prediction minus the bias). When we explicitly modeled the sensor bias, our (unbiased) predictions of the true system state better matched the measurements, an indication of a better estimate of the system state.

Finally, we tested the ability of the model to track non-steady-state behavior — in particular, the behavior of the system when the $CO_2$ supply is turned off during the shutdown process. Unfortunately, we only had one data set containing this transition, and thus we expect our parameters are still not tuned optimally. In addition, having only one such transition in our data, we report results on the same data that was used for training.

Fig. 5(e) shows a comparison between the predicted and measured output from pressure sensors $P_3$ and $P_4$, for two versions of the model. The first set of predictions, shown in solid lines, was calculated using our best estimates of the empirical parameters, including the membrane area (calculated from other parts of the data set) of 27.1. The second set of predictions, shown in dashed lines, was calculated using an earlier estimate of the membrane area of 31.6. While in the steady-state prior to timestep 220, the two predictions are equivalent as the differences were absorbed into the bias errors, in the transient part, the model with inaccurate parameters underpredicts the initial drop in pressure, and retains this error throughout the rest of the sequence.

[4]The liquid level sensor is very noisy, as splashing and bubbling from the dissolved $CO_2$ and from drops splashing from the condenser hit the sensor rod and create considerable noise in the sensor reading.

Fig. 5(f) presents the predictions of the correct model for the flows at $R_8$ (16) and $R_{12}$ (10), over a longer period of time. Initially, when the $CO_2$ supply was cut off, the flows dropped; however, gradually the CO and $CO_2$ in the system were vented and the only remaining gas was $H_2$. As the membrane presented less resistance to $H_2$ the flow rates started to go up. The model tracked this complex behavior surprisingly well.

## 7 Conclusions and Future Work

In this paper we address the problem of monitoring a large complex physical system — NASA's Reverse Water Gas Shift system — perhaps the largest and most complex hybrid DBN developed to date. This paper makes contributions both to the modeling and the monitoring of complex nonlinear systems. On the modeling side, we have shown how to model physical systems whose effects manifest themselves at dramatically different time scales, and that involve biased sensors, where the bias is state dependent and varies over time. On the monitoring side, we have presented a general framework for approximating nonlinear behavior using integration methods that extend the Unscented Filter, improving the accuracy of the approximation with minimal additional computation. Experimental results indicate that this approach is much faster and more reliable than particle filtering. More generally, we have demonstrated the feasibility of hybrid DBNs for monitoring a complex real-world physical system such as the RWGS using real data.

There are several interesting directions for future work. The tracking algorithms presented in this paper assume a known mode of operation, i.e., all the discrete variables are observed. Our long-term goal is to diagnose the RWGS when components fail. In order to track both the discrete and continuous state, we intend to combine the results presented in this paper with algorithms that handle hidden discrete events such as *Rao-Blackwellized Particle Filtering (RBPF)* [Doucet *et al.*, 2000] or the algorithms presented in [Lerner and Parr, 2001; Lerner *et al.*, 2000]. The speed of our algorithm (taking just 20ms to generate a Gaussian over all the state variables) is a promising indication that we can use these techniques for real-time fault diagnosis.

**Acknowledgements**

We are very grateful to Charlie Goodrich and the rest of the RWGS team at the Kennedy Space Center — Bill Larson, Clyde Parrish, Jon Whitlow, Curtis Ihlefeld, and Dan Keenan — for their tremendous help and support. We also thank Dan Clancy, Ronald Parr, and Stephen Boyd for useful suggestions and discussions. This research was supported by ONR Young Investigator (PECASE) under grant number N00014-99-1-0464, by ONR under the MURI program "Decision Making under Uncertainty", grant number N00014-00-1-0637, and by NASA under grant number NAG2-1337.